\documentclass[12pt]{article}
\usepackage{amsmath}
\usepackage{amsfonts} 
\usepackage{algorithm}
\usepackage{algorithmic}
\usepackage{times}
\usepackage{graphicx}
\usepackage{color}
\usepackage{multirow}
\usepackage{graphicx}
\usepackage{subcaption}
\usepackage[natbibapa]{apacite} 
\usepackage{rotating}
\usepackage{bbm}
\usepackage{latexsym}

\newcommand{\captionfonts}{\normalsize}

\makeatletter  
\long\def\@makecaption#1#2{%
  \vskip\abovecaptionskip
  \sbox\@tempboxa{{\captionfonts #1: #2}}%
  \ifdim \wd\@tempboxa >\hsize
    {\captionfonts #1: #2\par}
  \else
    \hbox to\hsize{\hfil\box\@tempboxa\hfil}%
  \fi
  \vskip\belowcaptionskip}
\makeatother   

\begin{document}
\hspace{13.9cm}

\vspace{20mm}

\begin{center}  
{\LARGE Optimizing Attention and Cognitive  Control Costs Using Temporally-Layered Architectures}
\end{center}

\ \\
{\bf \large Devdhar Patel$^{\displaystyle 1}$, Terrence Sejnowski$^{\displaystyle 2, \displaystyle 3,\displaystyle 4}$, Hava Siegelmann$^{\displaystyle 1}$}\\
{$^{\displaystyle 1}$
Manning College of Information and Computer Science, University of Massachusetts, Amherst, MA 01003, USA}\\
{$^{\displaystyle 2}$Computational Neurobiology Laboratory, The Salk Institute for Biological Studies, 10010 North Torrey Pines Road,
La Jolla, California 92037, USA}\\
{$^{\displaystyle 3}$Institute for Neural Computation, University of California San Diego, La Jolla, California 92093, USA}\\
{$^{\displaystyle 4}$Department of Neurobiology, University of California San Diego, La Jolla, California 92093, USA}
%


\thispagestyle{empty}
\markboth{}{NC instructions}
\ \vspace{-3mm}\\
%
\begin{center} {\bf Abstract} \end{center}
The current reinforcement learning framework focuses exclusively on performance, often at the expense of efficiency. In contrast, biological control achieves remarkable performance while also optimizing computational energy expenditure and decision frequency. We propose a Decision Bounded Markov Decision Process (DB-MDP), that constrains the number of decisions and computational energy available to agents in reinforcement learning environments. Our experiments demonstrate that existing reinforcement learning algorithms struggle within this framework, leading to either failure or suboptimal performance. To address this, we introduce a biologically-inspired, Temporally Layered Architecture (TLA), enabling agents to manage computational costs through two layers with distinct time scales and energy requirements. TLA achieves optimal performance in decision-bounded environments and in continuous control environments, it matches state-of-the-art performance while utilizing a fraction of the compute cost. Compared to current reinforcement learning algorithms that solely prioritize performance, our approach significantly lowers computational energy expenditure while maintaining performance.  These findings establish a benchmark and pave the way for future research on energy and time-aware control.

\section{Introduction} \label{intro}

Deep Reinforcement Learning (DRL) has demonstrated a remarkable capacity for learning control policies \citep{fujimoto2018addressing, haarnoja2018soft, Mnih2015HumanlevelCT}. However, existing efforts solely focus on maximizing predefined environmental rewards at a constant decision-making frequency. This approach contrasts with biological control, where organisms adjust their behavior and energy expenditure based on environmental demands. This adaptive method allows for more efficient resource usage and can enhance performance in complex, fluctuating environments where defining reward functions can be challenging and optimizing only the reward function can lead to unexpected behavior. 


Biological control is significantly more efficient, flexible, and effective compared to current artificial control methods despite severe time delays \citep{More2018ScalingOS}, slow rates of information transmission, and slow response times \citep{jain2015comparative}. These limitations can be ameliorated by integrating the capabilities of diverse components \citep{Nakahira2021DiversityenabledSS},
distributing control over many layers \citep{li2023internal}, and by incorporating multiple adaptive response times into these layers, as explored in this study. 


State-of-the-art reinforcement learning (RL) algorithms currently lack the ability to adapt their time step size \citep{haarnoja2018soft, fujimoto2018addressing, schulman2017proximal}. Typically, a constant time step is selected to avoid the problem of continuous optimization of the timestep in every state. As a result, RL agents prefer extremely fast frequencies. However, different environments have different temporal contexts, each requiring a different time step size for an optimal performance-energy trade-off. Moreover, even within the same environment, the optimal time step can change. For example, when moving from a relatively safe and predictable state to an unpredictable state, the timestep should decrease. Ideally, as the state transitions become more predictable with training and repetition, less sensing should be required, and larger time steps become feasible. A fixed time step cannot effectively handle dynamically changing environments and is the cause of many failures in current algorithms.

The RL framework models the environment as a Markov Decision Process (MDP). For continuous environments, an agent acting at a fixed frequency must operate at least as fast as the situation that requires the fastest response. Such fast response speed means that the agent divides an episode into more states, resulting in a longer task horizon. This can decrease action-value propagation and, in turn, slow convergence to optimal performance \citep{McGovern1997RolesOM}. 
 Similarly, the temporal density of the reward (reward/state transitions per unit time) decreases as the agent becomes faster since the total state transitions increase, leading to an increase in the difficulty of the RL problem. 
 This is especially true for environments with sparse rewards, where a faster agent would have to explore more zero-reward (uninteresting) state transitions before finally reaching the goal or failure state that provides a non-zero reward. The mountain-car problem is one such example of this scenario \citep{Moore90efficientmemory-based}. 
  
A faster response time also means that the agent processes more inputs per unit time and needs faster actuation, both of which require more energy. In energy-constrained agents like robots, the agent's response speed can have a significant impact on total energy consumption. It is worth noting that DRL algorithms rely on experience replay memory during learning \citep{Mnih2015HumanlevelCT}, and that a faster response time would result in the creation of more memories per unit time. Consequently, a small memory size would also bottleneck the performance of fast-acting RL. 


We primarily focus here on the effect of step size on energy conservation and performance. To investigate this formally, we introduce the Decision Bounded-MDP (DB-MDP), an extension of the MDP that constrains the number of decisions the agent can make in each episode. This constraint motivates the agent to conserve its decision-making energy expenditure. Fixed time-step RL will yield suboptimal solutions or even fail completely on DB-MDP.

To design an agent that can handle such environments, we take inspiration from biological systems. The design of the brain enables it to use various contexts, such as safety, energy availability, and performance-energy trade-off coefficient, to modulate its response time, ensuring accurate responses in both familiar and unfamiliar environments. This design allows for energy conservation in predictable situations, where slower reactions are acceptable while allowing for faster reactions in unpredictable situations. Recent work has shown that the brain uses distributed control to allow multiple independent systems to process the environment and react accurately \citep{Nakahira2021DiversityenabledSS}, building on the history of research on the speed/accuracy trade-off \citep{heitz2014speed}. This distributed control enables multiple layers of biological neural networks, with different delays and speeds, to activate and control muscle groups for executing complex behaviors. As a result, the brain and central nervous system can trade off between speed and accuracy, depending on the situation's demands.

\begin{figure}
  \centering
  \includegraphics[width=13cm]{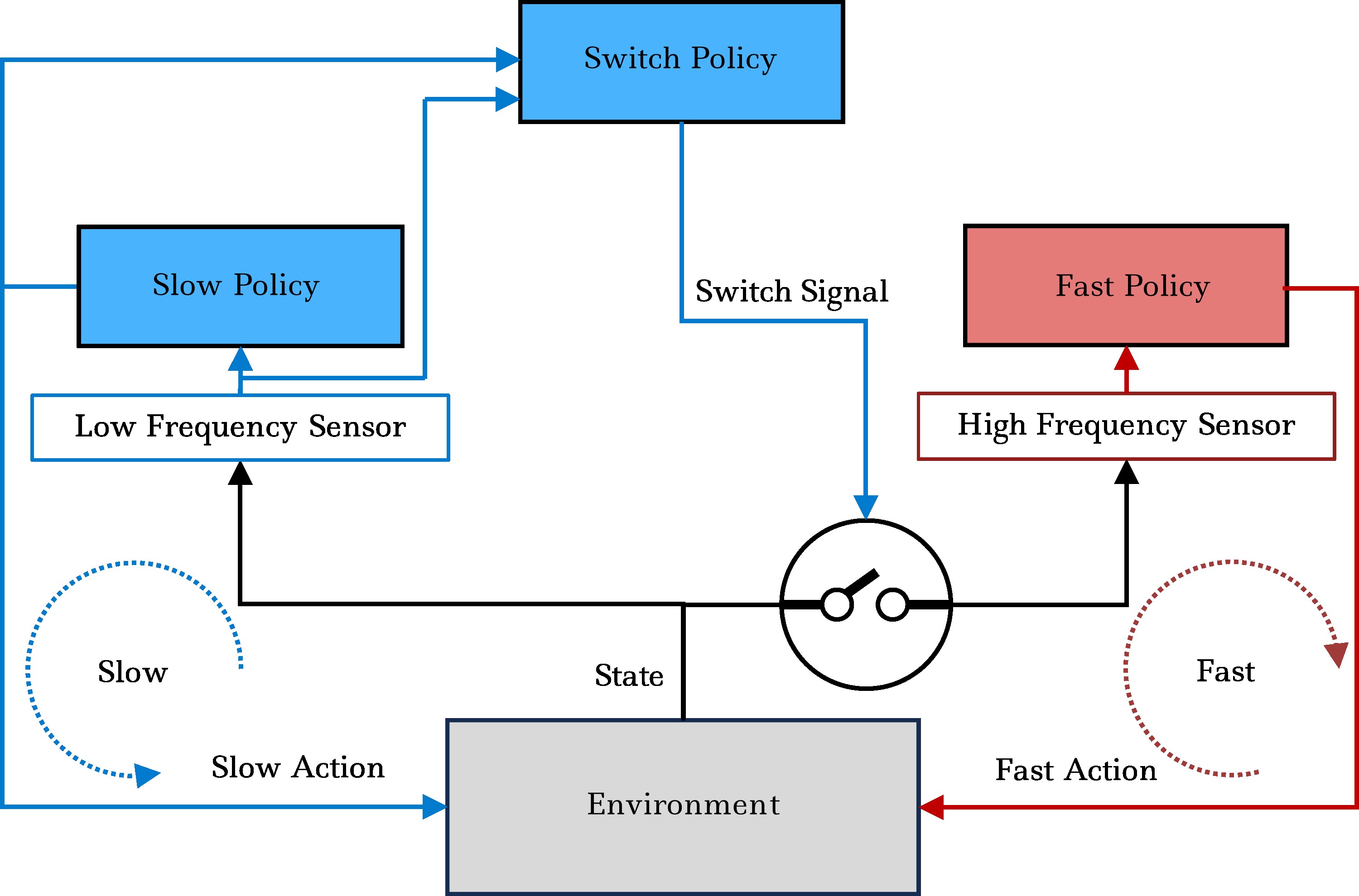}
  \caption{The Temporally Layered Architecture (TLA) comprises two layers: the Slow policy (blue) and the Fast policy (red). The switch policy can activate or deactivate the Fast policy, thus switching between the two layers. The reward given to each network is augmented differently with the energy and consistency penalty, which forces the overall policy to learn temporal abstractions from performance and energy-based contexts.}
  \label{TLA_fig}
\end{figure}

Inspired by the biological design, we propose Temporally Layered Architecture (TLA) (Fig. \ref{TLA_fig}): a reinforcement learning architecture that layers two different policy networks with different frequencies, allowing the RL agent to adapt its response frequency online by using their combination. To switch between the two policies optimally, we introduce a switch policy that is also trained using reinforcement learning. Training two different policies together to act on the same environment (albeit at different times) is a challenging multi-agent task with no communication between the policies. However, we introduce an algorithm that can not only train all three networks (Fast, Slow and Switch) together but also demonstrates competitive sample efficiency compared to single policy algorithms. To aid with the parallel training of the two networks, we introduce two different intrinsic reward penalties: an energy penalty and a consistency penalty. The goal of the energy penalty is to encourage the use of the slow layer, aiding learning by adding a constraint that reduces the number of optimal policies to the most efficient ones. The consistency penalty is an intrinsic reward that encodes the inconsistencies between the actions picked by the slow and fast policies, enabling the policies to learn from each other by mimicking behavior. 
\\
\textbf{Summary}:
\begin{enumerate}
    \item DB-MDPs extend the traditional MDP framework to include decision making constraints. Traditional reinforcement learning algorithms are limited in their ability to explore and navigate DB-MDPs. 
    \item A novel biologically-inspired Temporally Layered Architecture (TLA) allows each layer to focus on a different temporal context and to navigate the entire DB-MDP more effectively than classical RL agents.
    \item  The two layers of the TLA can be simultaneously trained with an efficient learning algorithm.
    

    \item We empirically test decision-bounded environments for both tabular and parameterized policy on gridworld and continuous control environments and demonstrate that TLA succeeds where the state-of-the-art algorithms fail.
    
    \item Empirical results on eight continuous control tasks 
    demonstrate that TLA is able to achieve state-of-the-art performance with less computational cost and fewer decisions.
    
\end{enumerate}

\section{Background}
Our novel control architecture combining multiple controllers with different response is relevant to several sub-fields of AI: 

\subsection{Continuous Control}
Continuous control refers to tasks that involve continuous actions. Compared to discrete control, exploration and learning for continuous control is more difficult and often requires a very fast response frequency. Additionally, continuous control agents often need to control multiple joints and actuators.

We use the state-of-the-art twin-delayed deterministic policy gradient (TD3) algorithm for continuous control \citep{fujimoto2018addressing}. TD3 learns two Q-functions (critics) and uses the pessimistic value of the two for training a policy that is updated less frequently than the critics. 
Because TLA does not depend on the RL training algorithm, it can easily accommodate different training algorithms. For example, for gridworld environments, TLA is implemented using Q-learning on tabular policies.

\subsection{Action repetition and frame skipping}\label{AR}
 Reinforcement learning with a sequence of actions is challenging since the number of possible action sequences of length $l$ is exponential in $l$. As a result, research in this area focuses on pruning the possible number of actions and states \citep{Hansen1996ReinforcementLF, Tan1991CostSensitiveRL, McCallum1996ReinforcementLW}. To avoid the exponential number of action sequences, some works have restricted the action sequences to repeating a single action. The number of actions is therefore, linear in the number of time steps \citep{Buckland1993TransitionPD, Kalyanakrishnan2021AnAO, Srinivas2017DynamicAR, Biedenkapp2021TempoRLLW, Sharma2017LearningTR}. Frame-skipping and action repetition have been used as a form of partial open-loop control, where the agent selects an action to be repeatedly executed without considering the intermediate states. TempoRL, introduced by \citet{Biedenkapp2021TempoRLLW}, learns an additional action-repetition policy that decides on the number of time steps to repeat a chosen action. This approach can lead to faster learning and reduce the number of action decision points during an episode. We use their approach as one of our benchmarks.

In their analysis of macro-actions, \citet{McGovern1997RolesOM} identified two advantages: improved exploration and faster learning due to a reduced task horizon. Empirical evidence from \citet{Randlv1998LearningMI} shows that macro-actions also significantly reduce training time. Additionally, \citet{Braylan2015FrameSI} showed that increasing the number of frames skipped can significantly improve the performance of the DQN algorithm \citep{Mnih2015HumanlevelCT} on some Atari games.

However, these approaches require a predictable environment so that an action can be repeated safely without supervision. Furthermore, these approaches often require additional hyperparameter search to find the best frame-skip parameter. In contrast, our approach can easily adapt its time step size to changing conditions and the current predictability of the environment. In addition, it is less sensitive to the frame-skip parameter as it can switch between networks of different time steps.

In a similar vein, \citet{Yu2021TAACTA} demonstrated a closed-loop temporal abstraction method in the continuous domain using an act-or-repeat decision after the action is picked, thus increasing action repetition. However, their approach requires two forward passes of the critic in addition to the actor and decision networks, as it uses the state-action value from the critic even after training. Thus, their approach does not reduce the number of decisions and is ill-suited for the DB-MDP problems.
Our approach (TLA) focuses on reducing the number of decisions and compute costs while increasing action repetition making it well suited for bounded decision environments.

\subsection{Residual and Layered RL}
Recently, \citet{Jacq2022LazyMDPsTI} proposed Lazy-MDPs where the RL agent is trained on top of a suboptimal base policy to act only when needed while deferring the rest of the actions to the base policy. They demonstrated that this approach makes the RL agent more interpretable as the states in which the agent chooses to act are deemed important. Similarly, for continuous environments, residual RL approaches learn a residual policy over a suboptimal base policy so that the final action is the addition of both actions \citep{Silver2018ResidualPL, Johannink2019ResidualRL}. Residual RL approaches have demonstrated better performance and faster training. Our approach is related to the residual approach, where a faster-frequency network is trained together with a slower-frequency base network to gain the benefits of macro-actions and residual learning. However, unlike the residual approach, the final action for TLA is exclusively picked by a single network. While residual approaches rely on a pre-trained base policy, the TLA demonstrates that both layers can be trained together. This is significant since TLA does not require a pre-trained policy and yet can train both layers from scratch with competitive learning speeds to single-layered RL.

\subsection{Options Framework}
The options framework \citep{Precup2000TemporalAI} is a common framework for temporal abstraction in RL. Options are defined as 3-tuples $\langle \mathcal{I}, \pi, \beta \rangle$. Where $\mathcal{I}$ is the set of initiation states which defines in which states the option can start; $\pi$ is the option policy that is followed for the duration of the option; and $\beta$ defines the probability of option termination in any given state. Options require prior knowledge about the environment to be defined. However, recent work has demonstrated that options that are automatically discovered by using the successor representation \citep{Machado2021TemporalAI} or the connectedness graph \citep{Chaganty2012LearningIA} can help improve exploration and thus learning. In a similar vein of research, \citet{Dabney2021TemporallyExtendedE} demonstrated that temporally extended actions improve exploration. Our work takes advantage of this phenomenon by layering the fast network and slow network to gain the exploration benefit of extended actions along with the precision of the fast network. 
In TLA, both or either of the fast and the slow networks can be formulated as options policy. Where the slow network encodes stateless options for open loop control while the switch network can learn the initiation states $\mathcal{I}$ and the option termination probability $\beta$.

\subsection{Multi-Agent Reinforcement Learning and Non-Stationarity}

Multi-agent Reinforcement Learning (MARL) is an open problem with many challenges \citep{zhang2021multi}. One of the main difficulties when training multiple agents is dealing with non-stationary environments \citep{Padakandla2020ASO}. In an environment where multiple agents interact during training, the transition function for each agent is not constant because the outcome depends on the joint action of all the agents. As a result, traditional reinforcement learning approaches based on the assumption that the environment can be modeled as a stationary MDP- fail to solve MARL tasks.

TLA is a unique cooperative MARL task in which all agents learn to control the same body together with limited information shared between the agents. In cooperative settings, many strategies have been proposed to train agents together \citep{Oroojlooyjadid2019ARO}. However, uniquely in our approach, we find that introducing energy constraints using differential intrinsic rewards induces cooperation and stable learning.

\section{Decision Bounded Markov Decision Process}

 The standard reinforcement learning setting \citep{sutton2018reinforcement} involves an agent that can take actions to cause state transitions in the environment, and in the process gain rewards. The environment is represented as a Markov Decision Process (MDP) \citep{puterman1990markov}. The goal of the agent is to maximize the amount of reward it gets during a single episode. Reinforcement learning problems are defined as typically defined as a 6-tuple $(\mathcal{S},\mathcal{A}, p, R, d_0,\gamma)$. Where:
\begin{itemize}
    \item $\mathcal{S}$ : set of all possible states in the environment
    \item $\mathcal{A}$: set of all possible actions the agent can take
    \item $p$: Transition function that defines how the environment changes. This is hidden from the agent. $p: \mathcal{S}\times\mathcal{A}\times\mathcal{S}\rightarrow[0,1]$
    \item $R$: Reward function. Typically hidden from the agent. $R: \mathcal{S}\times\mathcal{A}\rightarrow\mathbb{R}$
    \item $d_0$: Initial state distribution. $d_0: \mathcal{S} \rightarrow [0,1]$
    \item $\gamma$: discount factor
\end{itemize}

The agent is characterized by a policy $\pi: \mathcal{S}\times \mathcal{A} \rightarrow [0,1]$ The objective of the agent is to learn the optimal policy $\pi^8$, that maximises the expected sum of discounted rewards:
\begin{equation}
    J(\pi) := \mathbb{E}\left[\sum_{t=0}^{\infty}\gamma^t R_t | \pi\right] := \mathbb{E}\left[G_t | \pi\right]
\end{equation}
where $R_t$ is the reward at time $t$, and $G_t$ is the return from time $t$. 

 Here, the agent and the environment are defined as two distinct entities that can interact solely through actions. However, this setting does not capture the internal state of the agent, that can also be modeled as an MDP. For example, when modeling a battery-powered robot, if two different robots start in the same state of a deterministic environment and take the same actions but their algorithms consume different amounts of energy at each step, the state of the environment will be the same, yet it will not reflect the differing remaining charge in their batteries. \citet{Hansen1996ReinforcementLF} proposed control setting where sensing incurs a cost, thus allowing the agent to perform a sequence of actions to reduce the sensing cost. However, similar to the MDP, it does not capture the computation cost, as algorithms that maintain an accurate model of the environment can easily reduce their sensing cost by increasing the computation cost.

 \begin{figure}[H]
    \centering
    \includegraphics[width=1\linewidth]{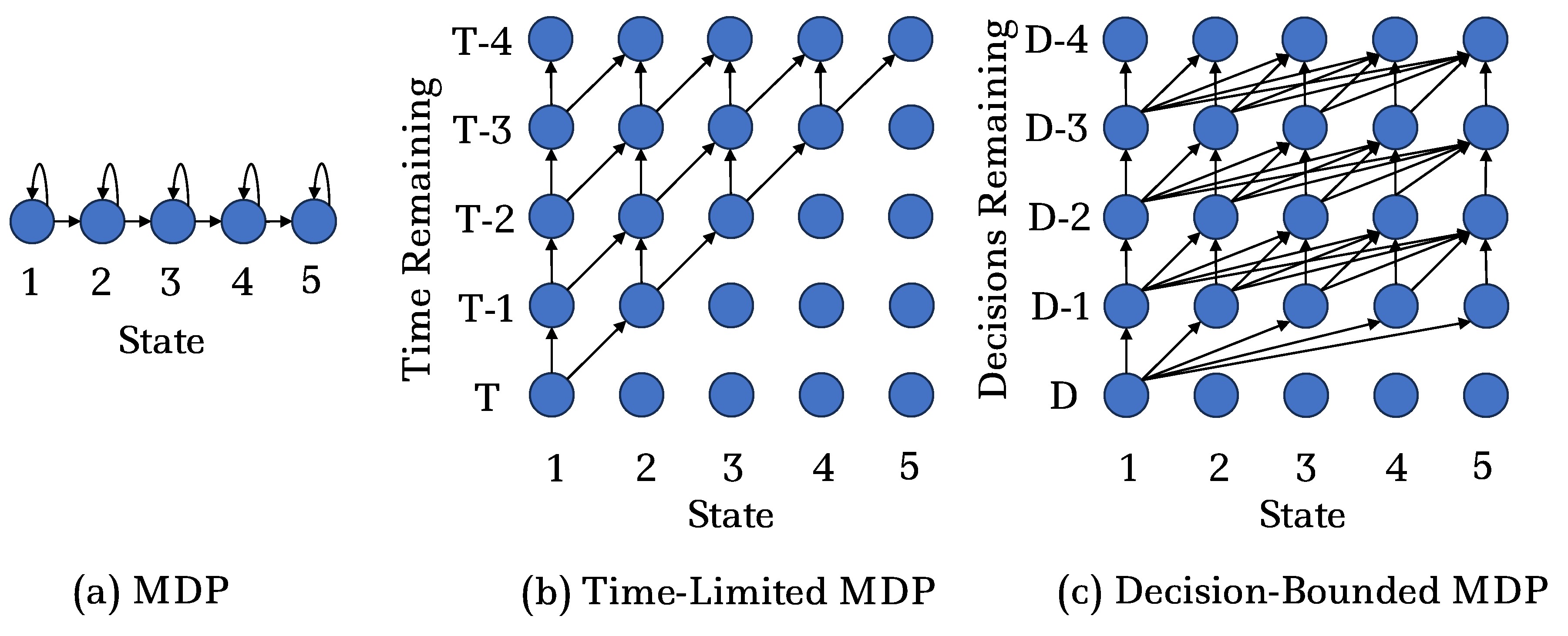}
    \caption{\textbf{(a):} A simple MDP with $S=5$ states. Each state has two actions, one that leads to the next state and one that results in the same state change.\textbf{ (b):} Time-limited MDP: In the time-limited MDP setting, there is an additional limit on the amount of time available ($T$). The MDP thus is expanded to $S \times T$ states.\textbf{ Right:} Decision-Bounded MDP: In Decision-Bounded MDP, the number of decisions are limited. However, a single decision can result in multiple planned actions. Similar to time-limited MDP, there $S \times D$ states where $D$ is the number of available decisions. However, a larger part of the MDP is reachable if the agent is able to take multiple actions per decisions, resulting in cognitive cost reduction.}
    \label{fig:MDPs}
\end{figure}

Thus, to capture this complexity, we introduce Decision-Bounded MDP (DB-MDP) that puts a bound on the number of decisions that can be taken by the agent in each episode. Decisions are defined as a sequence of $n$ actions where $n \in \mathbb{N}$.  We note that in time-limited tasks, simple agents that make one decision at each step ($n=1$) can be modeled as time-dependent MDPs \citep{pardo2018time}. Time-dependent MDPs can be thought of as a stack of $T$ time-independent MDPs followed by a terminal state. Therefore, at each time step, the actions result in transitioning to the next state in the MDP in the next stack. However, the reachable states in the next stack are the same as the states reachable from the current state in the original MDP.

However, when agents can plan a sequence of actions or repeat the same action ($n\geq 1$), it is possible to transition to states that are not adjacent in the original MDP thus preserving the number of remaining decisions (Figure \ref{fig:MDPs}). Thus, we can see that DB-MDPs are an extension of time-limited MDPs that dissociate the decisions from time. 

Decisions are not formally defined in the RL framework, yet they play an integral role in control. Typically, decision-making incurs a cognitive cost, and commonly repeated actions become more efficient over time in the brain as fewer decisions are required\citep{wiestler2013skill}. Thus, tasks like walking, while complex, do not require many decisions or much cognitive load. Formally, in this work, we define a decision as each time a single state is taken as an input to produce one or more actions. We note that this does not fully capture the cognitive energy expended behind each decision, which may be variable. However, most of current state-of-the-art algorithms incur the same amount of cognitive energy regardless of the complexity of the decision. Thus, decisions are directly proportional to the cognitive energy expended in such cases. Additionally, decisions also measure the amount of time between each forward pass. Since more decisions also mean that the agent has less processing time between each decision. Finally, decisions also captures the predictive power of each algorithm. Fewer decisions also mean that the agent can predict a longer action sequence. Thus, complementary to performance, decision is a versatile metric for the overall "goodness" of the algorithm.

\section{Temporally Layered Architecture}

This section discusses various methods for achieving temporal adaptivity in reinforcement learning. We introduce our novel architecture and its accompanying learning algorithm, which learns the two distinct temporal abstractions simultaneously and switches between them to optimize both performance and efficiency.

\subsection{Temporal Adaptivity}

In control tasks, different states require different levels of temporal attention. Some states are unpredictable, resulting in higher entropy for the transition function $p(s_{t+1},a_t,s_t)$. In these transitions, increased supervision is required to monitor and correct any undesired transitions so that the expected reward does not decrease after the action is taken. On the other hand, some states are predictable and have lower entropy for the transition function. In these states, since the outcome is expected and predictable, the agent can take more time before sampling input from the environment. The brain takes advantage of this phenomenon by reducing attention in familiar states while increasing it in unfamiliar or unpredictable states \citep{van2023effects, del2014oscillatory}. The primary benefit is to reduce the energy required for computation when it does not affect performance, thus increasing efficiency. 

In the context of RL, temporal adaptivity refers to changing the timestep $t$ based on the state. However, this is a non-trivial task because, for any given policy $\pi$, its expected sum of rewards $J(\pi)$ depends on the timestep $t$. This is because the reward gained from the environment for performing an action $a$ might change based on how long $(t)$ it is performed in the environment.

\subsection{Temporally Adaptive Reinforcement Learning}

A naive approach to adding temporal adaptivity is to treat each action-time step pair as a different action. The action space is augmented to include the time step, thus the policy is: $\pi: \mathcal{S} \times (\mathcal{A} \times \mathcal{T}) \rightarrow [0,1]$. However, this is undesirable as it makes the policy intractable due to the exponential increase in the possible number of actions. 

To overcome this issue, many approaches have been proposed that focus on increasing action repetition, as noted in \ref{AR}. Among them, the most successful in reducing the number of decisions is TempoRL  \citep{Biedenkapp2021TempoRLLW}.  TempoRL uses two networks: one to select an action and another to determine how long that action should be performed in the environment. However, they do not impose additional constraints or penalties to incentivize longer actions. Additionally, the actions are optimized for a single time step, so in situations where the optimal extended action differs from the optimal single-step action, TempoRL will not be able to learn the extended action.

\subsection{Temporally Layered Architecture (TLA)}

We draw inspiration from the brain and biological reflexes, which use multiple layers of computation with different latencies to enable temporal adaptivity. TLA has two layers, slow and fast, that learn two policies, each with a different step-size, $\pi^s$ and $\pi^f$, where $s$ and $f$ denote the slow and fast layers, respectively. The fast layer is similar to traditional RL agents and can observe and act at every time step, whereas the slow layer can only observe and act every $\tau$ time steps, where $\tau \geq 2$ and $\tau \in \mathbb{Z}$. Therefore for any $t$ mod $\tau$ that is equal to 0, the next action is sampled from the slow policy (Equation (2)), and the previous action is repeated otherwise. 

To switch between these two policies, we introduce a switch policy that decides whether to activate the fast network based on the state and the slow action. Therefore, at each time step:
\begin{align}
a^s_t = \begin{cases} a^s_{t-1} & \text{if} \ t\  \text{mod}\ \tau \neq 0 \\
\sim \pi^s(\cdot\mid s_t)& \text{otherwise} \end{cases}
\end{align}
\begin{equation}
    g_t = \begin{cases} 
         g_{t-1} &\text{if} \ t\  \text{mod}\ \tau \neq 0 \\
        \sim \mu^g(\cdot \mid s_t, a^s_t) & \text{otherwise}
    \end{cases}
\end{equation}

\begin{equation}
    a^f_t \sim \pi^f(\cdot\mid s_t)
\end{equation}

\begin{equation}
    a_t = a^s_t \cdot (1-g_t) + a^f_t \cdot g_t
\end{equation}
Where $s_t$ is the state at time $t$,  $a^s$ is the slow action, $a^f$ is the fast action, and $g \in \{0,1\}$ is the switch action. $\pi^s, \pi^f, \mu^g$ are the slow, fast, and switch policies respectively. Thus, the fast network is only activated when $g=1$.

 Thus, during an episode, the agent might utilize both layers depending on the context. The value of any given state is dependent on both the fast and slow policies, making it is difficult to simultaneously optimize both layers. Each layer is unaware of the other layer's policy and thus needs to navigate in a non-stationary environment. To aid training, experiences are added to the replay memory of both the slow and the fast networks whenever either network is activated. This is straightforward for the fast network, as it has $\tau$ experiences with the same action created whenever a slow action is chosen. However, the slow network can only observe its own actions. When the fast network is activated, we augment the slow reward with a consistency penalty that captures the difference between the slow and the fast actions, facilitating the sharing of information between them. Additionally, the rewards of the slow and switch networks are augmented by the energy penalty to incentivize slow actions. Hence, the final rewards for each network are as follows:

\begin{equation}
    R^f_t = R_t - g_t \cdot (|a^s_t - a^f_t|/a_{max}) \cdot j
\end{equation}
\begin{equation}
     R^s_t = R_t - p \cdot g_t -g_t\cdot(|a^s_t - a^f_t|/a_{max})  \cdot j
\end{equation}

Where $j$ is the consistency penalty parameter and $p$ is the energy penalty parameter that incentivizes slow actions. Thus, even though the action $a^s$ is not performed, it can affect the reward,  aiding the training of the slow network. $R^q_t$ is the reward provided to the fast network at each step while the slow and the switch network receive the sum of rewards for $\tau$ steps: $\sum_{k = t-\tau}^t R^l_k$. This formulation allows the use of a single hyperparameter $j=p$ for simpler environments with consistency and energy penalties. For multi-dimensional environments that already have an action magnitude penalty, $j$ is set to zero, and only $p$ needs to be searched. Additionally, for multi-dimensional environments, since $j$ is set to 0 when the fast network is activated, the slow action has no influence on the reward received or the next state. This increases the relative non-stationarity of the slow-network. To avoid this, we set the slow action to zero every time the fast action is picked for the environments with multiple action dimensions. Thus, all non-stationary memories in the replay memory of the slow network have zero action while all non-zero action memories will be stationary. 

The slow and fast networks are trained using the TD3 algorithm, while the switch policy is trained using Deep Q-learning \citep{Mnih2015HumanlevelCT}. For completeness, the pseudo-code and an alternate architecture diagram with reward penalties is presented in the appendix.

\section{Experiments}

We evaluate the performance of TLA on various environments including decision-bounded, gridworld, and continuous environments. Additionally, we open-source our code \footnote{https://github.com/dee0512/Temporally-Layered-Architecture} and we compare TLA to three different benchmark algorithms:

\begin{enumerate}
   \item \textbf{Base algorithm:} TLA can utilize any model-free algorithm. In this work, we use Q-learning for gridworld environments and TD3 for continuous control. Therefore, we also compare TLA to the original algorithm that utilizes a constant timestep. In environments without a decision bound, this algorithm provides an upper-bound on reward since it represents the state-of-the-art.  
    \item \textbf{Extended-action base algorithm:} Since the slow layer of TLA utilizes a larger timestep, we also evaluate the base algorithms with larger constant timesteps. This algorithm provides a lower bound on the decisions for TLA as it always picks the slow timestep.
    \item \textbf{TempoRL:} To our knowledge, we are the first to introduce an algorithm that optimizes the dual objectives of performance and decisions in continuous control environments. However, we provide the TempoRL algorithm as an additional comparison, as it utilizes action repetition as a means of further optimizing performance and in the process, reduces the number of decisions. 
\end{enumerate}

\subsection{Decision Bounded Environments}

\subsubsection*{Discrete Decision-Bounded Environment (Gridworld)}
\begin{figure}[H]
    \centering
    \includegraphics[width=10cm]{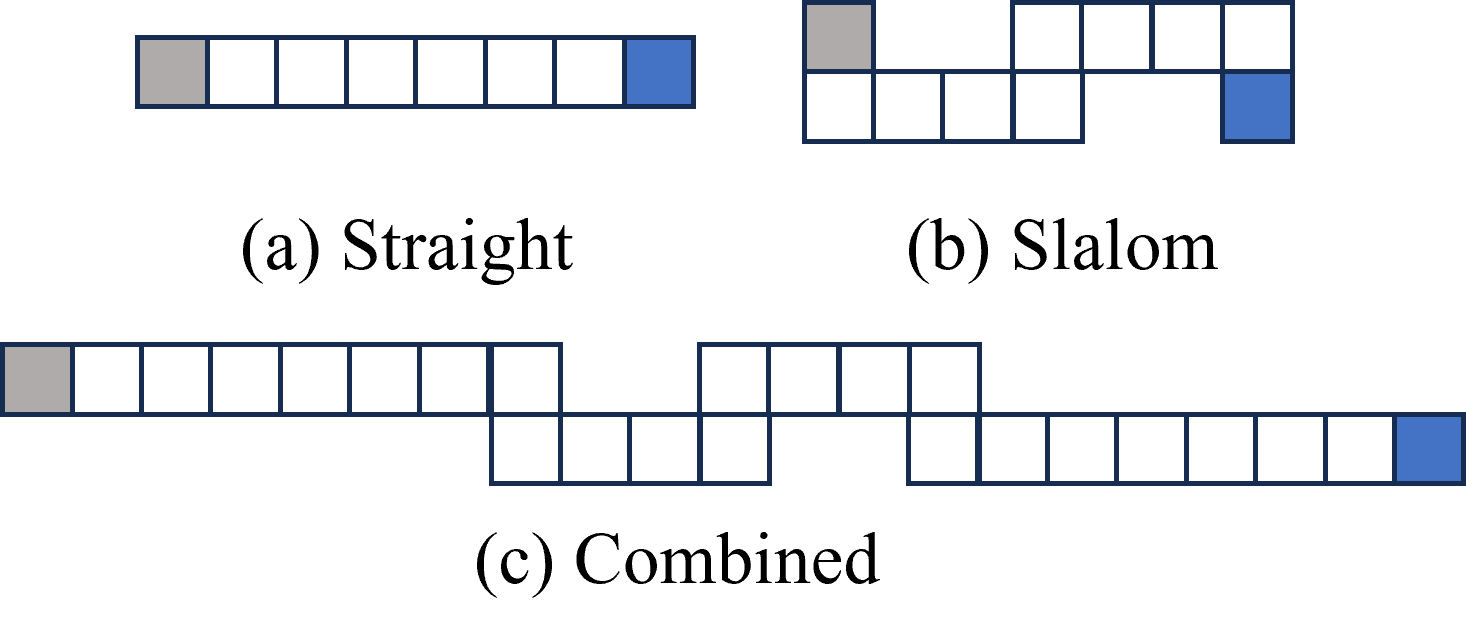}
    \caption{Gridworld Environments. The grey box represents the starting state and the blue box is the goal state.}
    \label{fig:gridworld}
\end{figure}

Since many aspects of reinforcement learning are linked to the time step, temporal adaptivity allows TLA to focus on multiple objectives and gain an advantage in performance, learning speed, and the number of decisions in decision-bounded environments. To demonstrate this, we introduce three different gridworld environments, each suited to a different optimal step-size: Straight, Slalom, and Combined (Figure \ref{fig:gridworld}). The Straight environment consists of a straight corridor of length 30. This environment can be easily solved by repeating a single action, thus it is easy to reduce the number of decisions in this environment. The Slalom environment consists of long horizontal corridors connected with short vertical turns where action repetition is suboptimal. Finally the Combined environment, combines the Straight and Slalom, so that the episode starts and ends with Straights and has Slalom in between. The environments are deterministic with four available actions to move in one of the four directions. At each environment transition, the agent receives a reward of -1. Upon reaching the goal, the episode ends. However, if the agent runs out of decisions before reaching the goal, the agent receives a large penalty of -50 reward for Straight and Slalom and -100 for combined. The choice of very large negative reward is chosen solely to distinguish between agents that run out of decisions and agents that do not. The number of decisions is limited to 15 for Straight and Slalom and 60 for combined.

We use Q-learning and tabular policies for evaluation and compare three different algorithms. Q-learning is a standard policy that takes one decision per time-step. Extended-action Q-learning repeats each action four times, thus reducing the number of decisions per unit of time by a factor of four. Finally, TLA can switch between taking one decision per time step and repeating an action four times. All values in the tabular policies are initialized to 0. Since in the tabular policy, the consistency penalty cannot be implemented, $j$ is set to 0 while $p$ is set to 1 (Eq. 6 and 7).

\begin{figure}
    \centering
    \includegraphics[width=1\linewidth]{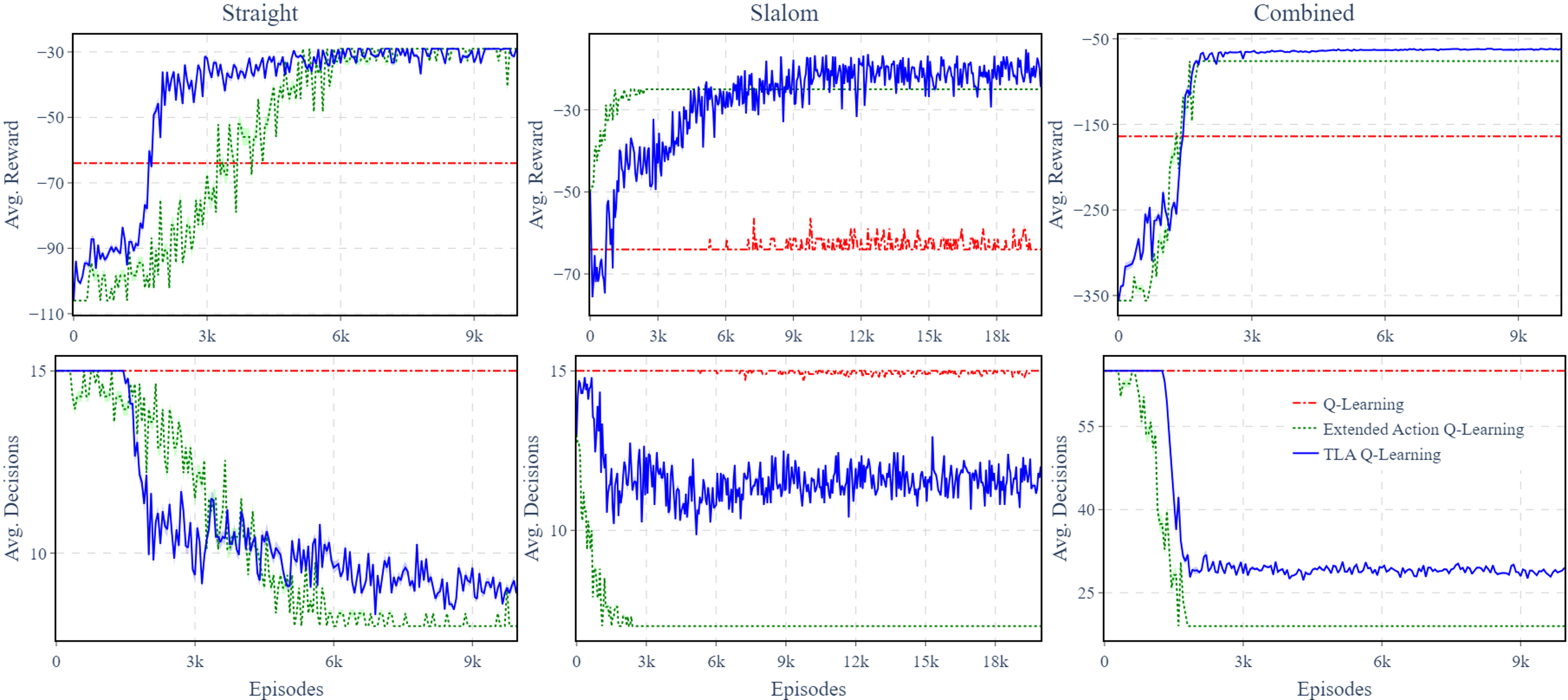}
    \caption{Decision Bounded Gridworld environments. TLA (blue) achieves the optimal performance with the fewest required decisions. All results are averaged over 20 trials. Top: Average reward vs. Episodes during training. Bottom: Decisions vs. Episodes during training.  }
    \label{fig:gridresults}
\end{figure}

We tested each policy on each environment over 20 independent trials. Fig. \ref{fig:gridresults} shows the average reward and average number of decisions vs. training episodes. When optimizing the dual goals of reward (performance) and decisions, priority is given to the reward to maintain competitive performance compared to decision-agnostic algorithms.

In the Straight environment, TLA quickly explores and reaches the goal and then optimizes the number of decisions. Curiously, it converges in performance faster than the extended action Q-learning even though the environment is ideal for action repetition. On the other hand, it is slow to converge towards the optimal decisions as it prioritizes performance over efficiency. Finally, Q-learning fails to solve the environment as there are fewer decisions allowed than the distance to the goal (Fig. \ref{fig:gridresults}). 

In the Slalom environment, the optimal path length is equal to the decision bound, thus it is possible for Q-learning to solve the environment. However, the decision bound makes exploration for Q-learning difficult. Extended-action Q-learning cannot reach optimal reward as it cannot navigate the sharp turns effectively. TLA is the only algorithm that reaches optimal reward. We see that TLA needs to take more decisions than extended action Q-learning. This is because it prioritizes reward over decisions. However, this behavior can be modulated by changing the energy penalty $p$. 

Similarly, in the Combined environment, optimal performance requires fast actions for the sharp turns and extended action elsewhere. In conclusion, TLA successfully switches between the two to achieve the perfect trade-off between performance and decisions so that it finds the optimal performance with the fewest possible decisions when acting at two timescales.

\subsubsection*{Continuous Control}
Next, we evaluate the performance and decisions of four different agents: TLA, TempoRL, TD3, and TD3 extended action (TD3-EA) on two decision-bound continuous action environments: LunarLanderContinuous-v2 and MountainCarContinuous-v0. We modified the Gym environments \citep{openai} to create decision-bounded environments. The number of decisions was constrained to 70 and 200 for the Lunar Lander and Mountain Car, respectively. Unlike the gridworld environments, there is no negative reward when the agent runs out of decisions. 

\begin{figure}
    \centering
    \includegraphics[width=1\linewidth]{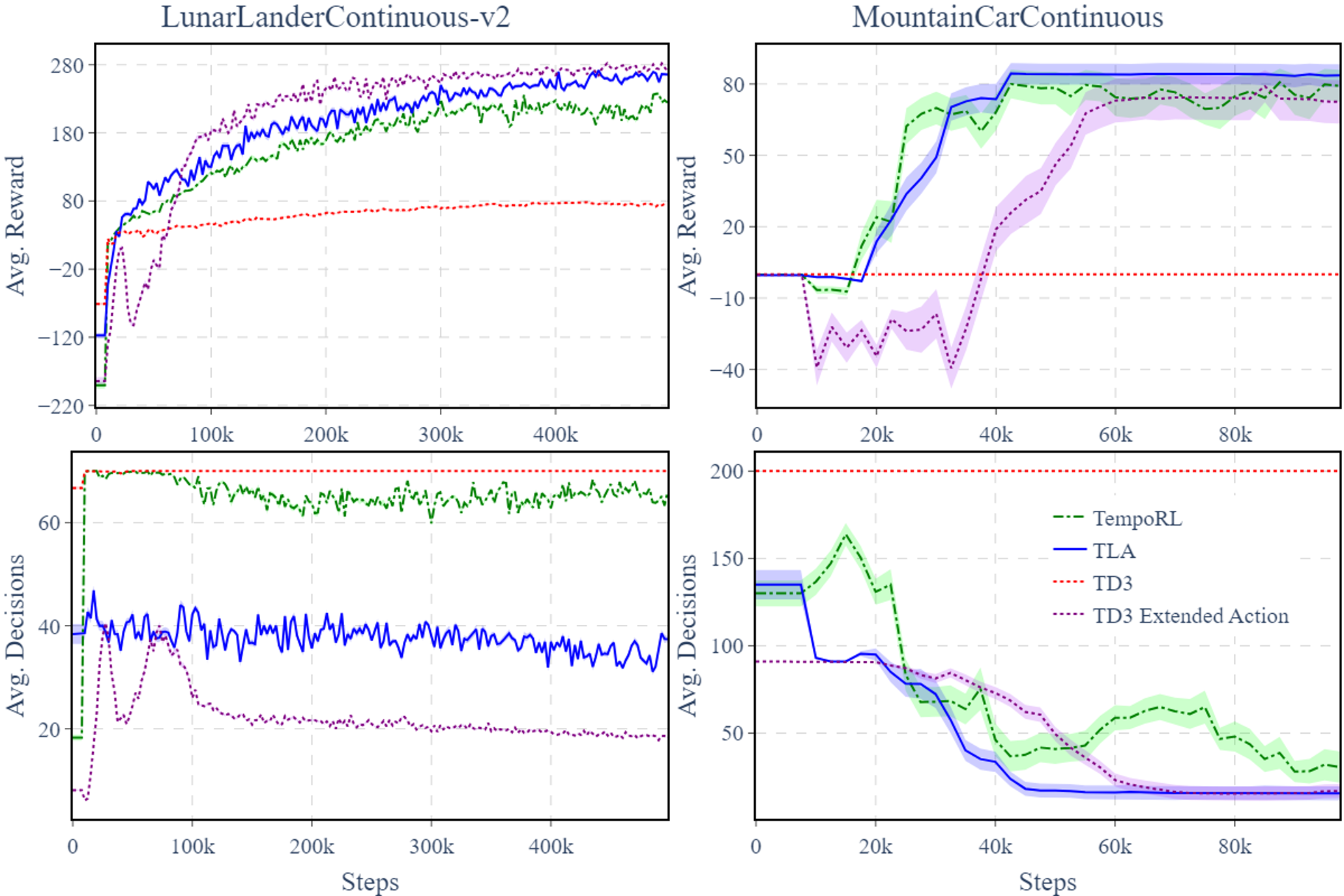}
    \caption{Decision Bounded Continuous Control environments. Top: average reward vs. training episodes. Bottom: Decisions vs. training episodes. All results are averaged over 10 trials. The shaded region represents standard error. Left: In the Lunar Lander environment, which is robust to action repetition, TD3 extended action (TD3-EA) shows superior performance. In this environment, it takes longer for TLA to converge towards optimal average reward, and thus decisions are not properly optimized during the training period as TLA prioritizes reward over decisions. Yet, TLA outperforms other algorithms and is able to successfully solve the environment under the decision constraint. Right: Due to the longer step-size, TLA, TD3-EA and TempoRL are able to successfully solve the Mountain Car task. However, TLA achieves better performance than TempoRL and TD3-EA while achieving the lower bound on decisions represented by TD3-EA.}
    \label{fig:ccdb}
\end{figure}

The maximum skip length for TempoRL, action repetition for TD3-EA, and $\tau$ for TLA are set to be the same for a fair comparison. The $\tau$ is set to 12 for Lunar Lander and 11 for Mountain Car. Figure \ref{fig:ccdb} shows the learning curves for rewards and decisions. All results are averaged over 10 trials. We find that the Lunar Lander environment is well-suited for extended actions. TLA, TempoRL, and TD3-EA achieve comparable performance on Lunar Lander. While TempoRL is not able to reduce the number of decisions effectively, TLA achieves the lowest decisions without forced extended action. We found that the Lunar Lander environment is uniquely robust to large timesteps and therefore, TD-EA achieves superior result. However, as we demonstrate in the following section, extended action algorithms are not suitable for most environments. 

TD3 cannot solve the Mountain Car environment due to inefficient exploration. We see that TLA is simultaneously able to optimize average reward and decisions, thus outperforming all algorithms. . While TempoRL is not able to optimize the number of decisions effectively, TLA achieves the fastest convergence to the highest performance with the fewest decisions. 

\subsection{Decision Unbounded Continuous Control Environments}

We evaluate TLA on a suite of 8 continuous control environments using the OpenAI gym library \citep{openai}: two classic control problems and six MuJoCo environments \citep{todorov2012mujoco}. We selected tasks on which TD3 has demonstrated state-of-the-art performance to highlight the inefficiency of the policies learned by current state-of-the-art algorithms. We set the time step of the fast-network to be equal to the default step size of the environment so that TLA has the same step size as TD3 and TempoRL. For the TempoRL algorithm, the max skip length $J$ was set to be equal to $\tau$, so the longest action repetition possible is the same for both TLA and TempoRL. Additionally, for reference, we also present the TD3-EA results. However, continuous control tasks are especially difficult to learn using extended actions, and we find that TLA utilizes both the fast and the slow layer for almost every task. Therefore, for a fair comparison, we set the time step of TD3-EA to be approximately equal to the average decisions-step size of TLA after training.

The algorithms' hyperparameters and neural network sizes were kept the same as in previous work \citep{fujimoto2018addressing}. The maximum training steps were set to 30,000 for the Pendulum-v1 environment and 100,000 for MountainCarContinuous-v0. The rest of the environments were trained until 1,000,000 steps. The initial exploration steps were set to 1,000 for Pendulum-v1, InvertedPendulum-v2, and InvertedDoublePendulum-v2; 10,000 for MountainCarContinuous-v0; and 20,000 for Hopper-v2, Walker2d-v2, Ant-v2, and HalfCheetah-v2. A complete list of hyperparameters is included in the appendix.

For each environment, a hyperparameter search for $\tau$ and $p$ was conducted over 5 random seeds. The final results presented are averaged over 10 random seeds. The hyperparameter search for $\tau$ was limited to a maximum of 11, and $p$ was evaluated over the range [0.1, 6]. Note that for different environments, the average reward per time step varies, and therefore, the optimal value of $p$ also varies with it. The environments with multidimensional actions (Hopper-v2, Walker2d-v2, Ant-v2, and HalfCheetah-v2) have a control cost included in their rewards, which is similar to the consistency penalty. Thus, for those environments, $j=0$. For the rest of the environments, $j=p$ for simplicity (Eq. 5 and 6). 

We note that learning extended actions for multi-dimensional actions is more difficult than for single-dimensional actions as it enforces the repetition syncing across all the dimensions. Therefore, we split the results into single action dimension and multiple action dimensions.


\subsubsection*{Single Action Dimension}

\begin{figure}[H]
    \centering
    \includegraphics[width=1.0\linewidth]{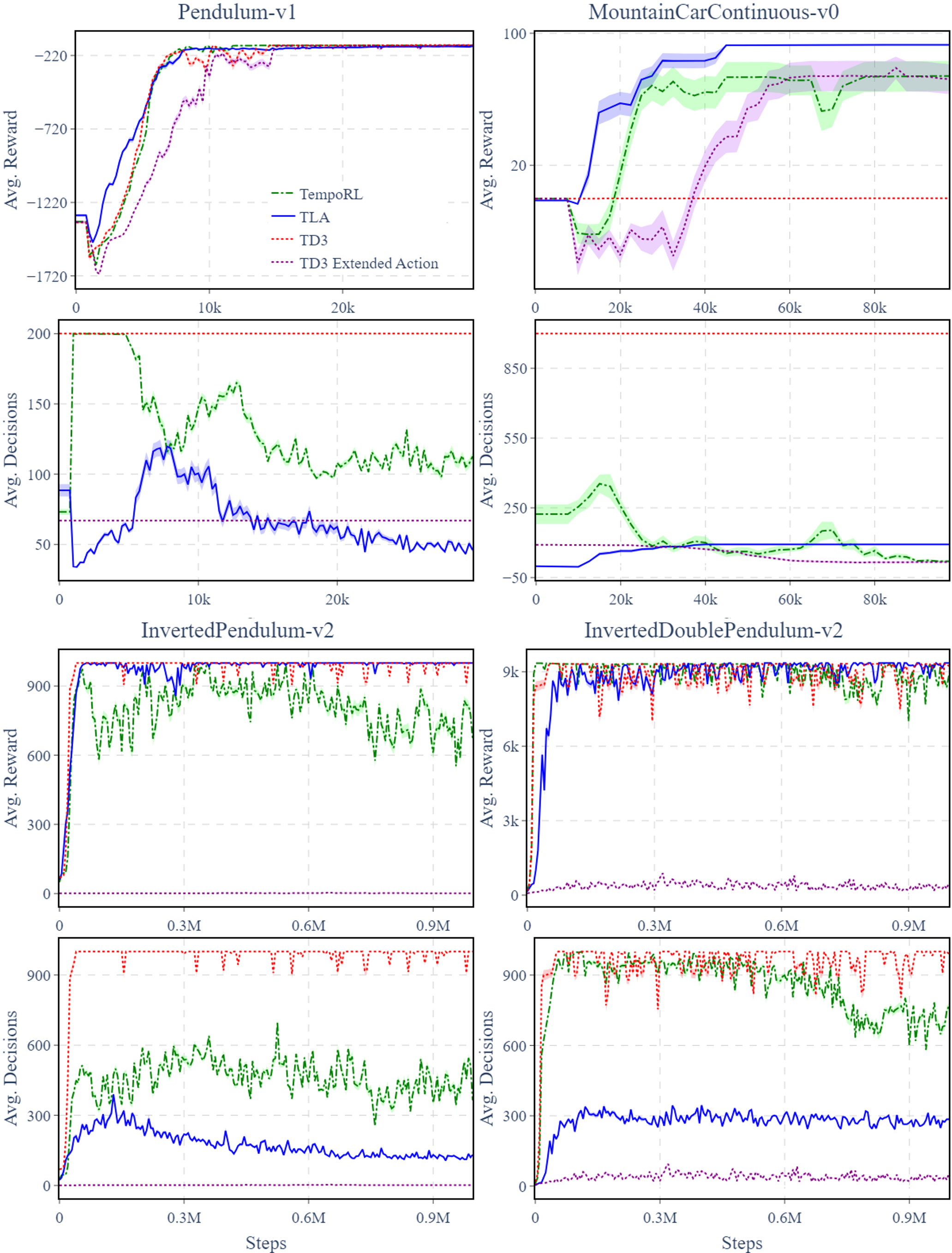}
    \caption{Average reward and average decisions during training. TLA (Blue) achieves state-of-the-art performance using a fraction of the decisions.}
    \label{fig:sd}
\end{figure}
We evaluate on four environments: Pendulum, Mountain Car Continuous, 
Inverted Pendulum, and Inverted Double Pendulum. Figure \ref{fig:sd} shows the average reward and decisions for the four different algorithms. We see that TLA outperforms all algorithms in all four environments (prioritiziing high reward, then lower decisions). On the Mountain Car environment, TLA utilizes more decisions than TempoRL and TD3-EA to achieve greater rewards. In the rest of the environments, TLA achieves optimal performance while achieving the lowest number of decisions. TD3-EA, on the other hand, fails to learn the Pendulum tasks, and thus results in very few decisions before the episode ends. We find that on the Inverted Pendulum task, after training, TLA almost never activates the fast network, resulting in an optimal policy with a timestep that is 10 times larger. Yet this policy cannot be learned by standard RL algorithms that only utilize a single layer, as evidenced by TD3-EA that has the same timestep.

\subsubsection*{Multiple Action Dimensions}

\begin{figure}
    \centering
    \includegraphics[width=1.0\linewidth]{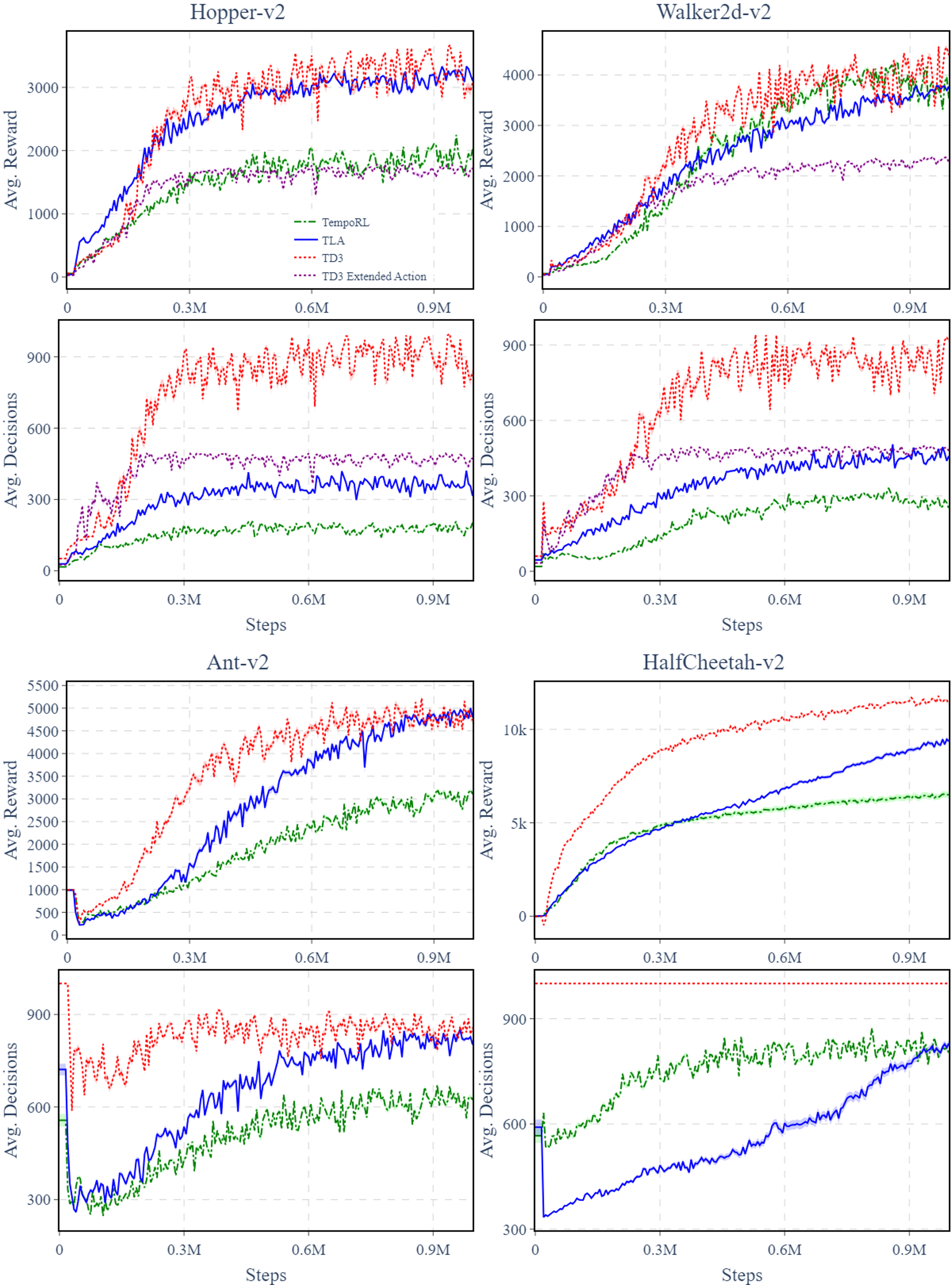}
    \caption{Average reward and average decisions during training on environments with multiple action dimensions. Multiple action dimensions are not well suited for macro actions that are action repetitions yet TLA (Blue) achieves comparable performance using a fraction of the decisions.}
    \label{fig:md}
\end{figure}

To demonstrate the scalability of TLA, we also present results on four difficult continuous control problems that are ill-suited for action repetition: Hopper, Walker2d, Ant, and Half Cheetah. In multidimensional environments, action repetition forces all actuators to repeat the same action for the same amount of time in a synchronous manner. This results in a suboptimal policy, as repetition syncing is almost never the optimal behavior.

Figure \ref{fig:md} presents the reward and decision learning curves for the four different algorithms. Surprisingly, TLA outperforms TD3 on the Hopper environment despite the challenge of repetition syncing. Additionally, in all environments tested, TLA achieves comparable performance with fewer decisions. On the other hand, TempoRL cannot scale to the difficult environments of Ant and Half Cheetah.

\subsection{Energy, Action-Repetition and Jerk}

As mentioned earlier, the number of decisions gives an overall picture of many important underlying metrics like cognitive cost, actuation cost, and reaction time. However, while generally it is beneficial to reduce the number of decisions, it results in different scaling of individual metrics for different algorithms. For example, it might be possible to employ a very accurate model of the environment and plan a sequence of actions using online planning to reduce the number of decisions, but this is a prohibitively costly strategy in terms of cognitive cost. 

Thus we measure four additional different metrics that are affected by reduced decisions and macro-actions:

\begin{enumerate}
    \item Computation Cost: The computation cost is dependent on the actual neural network architecture and the algorithm employed. Since the algorithms we evaluate have different architectures with different numbers of parameters, the cognitive cost can vary significantly. TLA has roughly three times more parameters since it utilizes three policies (fast, slow, and switch) and yet TLA demonstrates that it utilizes a fraction of the computation cost compared to other algorithms. On the other hand, while TempoRL reduces the number of decisions, it often has a higher computational cost. Figure \ref{fig:macs} demonstrates the average Multiply-Accumulate operations (MACs) per episode vs. training steps for all algorithms. 

    \begin{figure}
    \centering
    \includegraphics[width=1.0\linewidth]{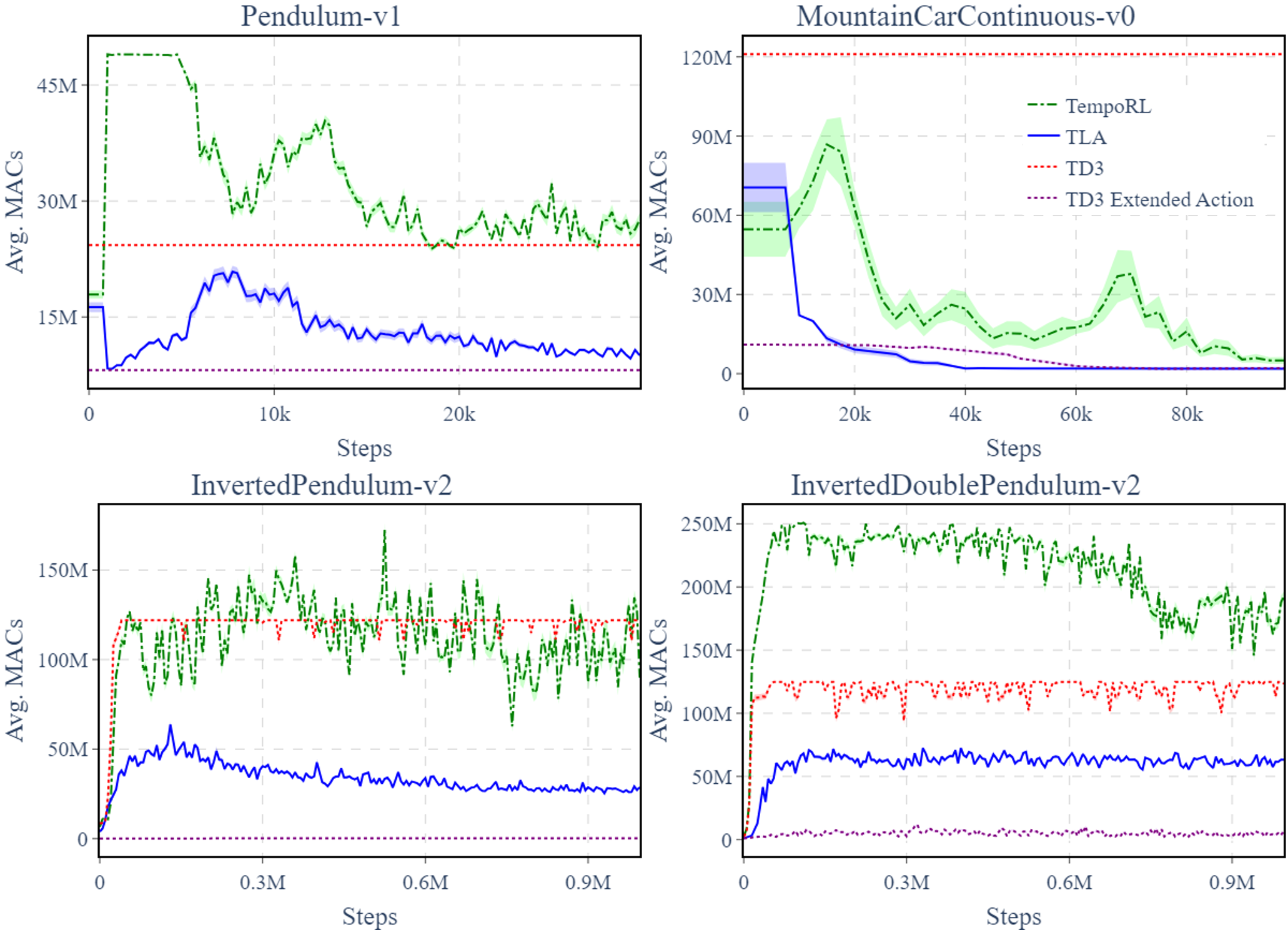}
    \caption{Average Multiply-Accumulate operations (MACs) vs. training steps: TLA (blue) uses only a fraction of the compute cost, even though it has roughly 3x the parameters of TD3 and 1.5x the parameters of TempoRL.}
    \label{fig:macs}
\end{figure}

    \item Action Repetition: In real-world tasks, there is often latency and communication cost involved between action selection and actuation. In such applications, increasing action repetition can reduce the amount of communication required since the same action can be repeated until a new action or directive is received. Therefore, we measure action repetition percentage as the average percentage of time steps in an episode where the previous action was repeated. For multi-dimensional actions, we calculate action repetition individually over each dimension before averaging, as each action dimension represents a different actuator that requires a separate channel of communication. Unsurprisingly, TLA and TempoRL have significantly higher action repetition across all environments. We provide detailed results in the Appendix. 

    \item Jerk: Additionally, the single most important metric in continuous control is jerk. The motions of the human body minimize jerk in their behavior to reduce joint stress and energy cost \citep{802610}. Thus, it is desirable to reduce jerk in the control task as it reduces energy expended during actuation and lowers the risk of damage or wear to the actuators \citep{tack2007relationship}. We measure jerk as the difference in action magnitude per time step, as each action represents the torque or force applied. We find that TLA reduces jerk in all but one environment. We provide detailed results in the Appendix.

    \item Area Under Curve (AUC): The area under the average reward training curve is used by previous works to measure the convergence speed of RL algorithms \citep{Biedenkapp2021TempoRLLW}. One benefit of using action repetition and macro-actions is that the agent can explore the environment more effectively and thus converge to optimal performance quickly. However, we find that this is not always the case in continuous action spaces. TempoRL demonstrated better AUC in discrete environments, yet when tested in continuous environments, it fails to converge faster than TD3. On the other hand TLA, despite training three different policies in parallel, is able to outperform TD3 in three of the eight tested environments. 

\end{enumerate}

\begin{table}[t]
\centering

\begin{tabular}{|l|l|l|l|}
    \hline
    Metric & TD3 & TempoRL & TLA (Ours) \\
    
    \hline \hline
    Avg. Reward & \textbf{7} & 3 & \textbf{7}\\
    Normalized AUC & \textbf{5} & 0 & 3 \\
    Action Repetition & 0 & 1 & \textbf{7} \\
    Jerk & 1 & 1 & \textbf{6} \\
    Decisions & 0 & 1 & \textbf{7} \\
    Compute Cost (MACs) & 2 & 1 & \textbf{5} \\
    \hline

\end{tabular}

\caption{Number of environments (out of 8) where the evaluated algorithm demonstrated the best performance on the given metric. For the Avg. Reward, all values within one standard deviation of the best performance are accepted. TLA demonstrates a well-rounded policies when compared to other algorithms.}
\label{table:ar}
\end{table}
 
We demonstrate that environments can have multiple solutions (optimal policies), and they are difficult to distinguish solely based on performance (average Reward). Table \ref{table:ar} summarizes the evaluation of all the algorithms on all the metrics tested. We note that TD3-EA is not incuded in this table since it is not a different algorithm, rather just TD3 with the timestep changed and it reaches acceptable performance on only 2 out of the 8 environments. For each RL algorithm, since the first task is to find the optimal policy, we first measure the average reward. To reduce the effect of randomness, we accept all values that are within one standard deviation of the best reward. Additionally, for the Inverted Double Pendulum environment, we accept all solutions since it is designed to have a very high reward while the optimized policies display a very low standard deviation (detailed results in the Appendix). 

Finally, for every environment where the algorithm reaches a competitive solution as described above, we further test it on six different metrics: AUC, action repetition, jerk, decisions, MACs. We find that TLA reaches a competitive performance in 7 out of the 8 environments (all except Walker2d). Furthermore, it also demonstrates that the policies learned by TLA are well-rounded and demonstrate superior performance when measured on other metrics. Importantly, it demonstrates higher action repetition and fewer decisions on all 7 environments. Additionally, it demonstrates lower jerk in all but 1 environment and lower compute cost in all environment except the Ant and Half Cheetah, where it suffers from repetition syncing.

We provide more detailed results on each metric in the Appendix.

\section{Conclusion}

This paper focuses on a new biologically-inspired methods to make RL more practical in realistic environments. Learning enhances layered control architectures \citep{li2023internal} and optimizes the speed-accuracy tradeoffs found in biological control systems \citep{Nakahira2021DiversityenabledSS}. We first introduce the Decision-Bounded MDPs and demonstrate that state-of-the-art RL approaches are unable to optimize their strategies, and frequently they completely fail.  Our Temporally Layered Architecture (TLA) is able to change the amount of decisions and compute based on realistic needs and thus can effectively solve problems in decision-bounded environments. We then tested the TLA in decision-unbounded environments and showed its superiority there as well: TLA is able to find more efficient solutions due to its temporal adaptivity. Additionally, we show how optimizing the dual objectives of performance and energy results in a reduced jerk, resulting in more natural and safe control. 

Our new architecture, the TLA, achieves temporal awareness by allowing the agent to choose between two different policies that make up its two layers. The slow layer allows TLA to plan a sequence of actions without intermediate sampling from the environment. Thus, the slow layer operates at a higher latency using a partially open-loop control where the next action in the sequence of actions does not depend on the next state sampled of the environment. On the other hand, the fast action acts similarly to the traditional reinforcement agent and is reflexive (reactive) in nature. It is a closed-loop system where each action depends on the state resulting from the previous action. A third network helps in achieving the switch between these two, and together, the layered system mimics the biological control achieved by the spinal cord and the brain \citep{Weiler2019SpinalSR}. Similar parallel pathways are also present inside the brain that allows it to change between planning and reflex depending on the situation \citep{Nakahira2021DiversityenabledSS}. A look at the architecture of the brain demonstrates that it is designed in this layered manner. The brain has connections to the superior colliculus from almost every other area \citep{Harting1992Corticotectal}. The superior colliculus is a site of sensorimotor integration responsible for motor control. This might allow the brain to spend variable amounts of compute resources and change its latency while picking actions. We note that this idea is also related to the idea of early exits in deep neural networks \citep{patel2022quicknets, scardapane2020should}, however, our approach proves that it is better suited for control where the difference in compute and latency has a significant impact. Furthermore, recent work has also shown evidence for parallel multi-timescale learning, similar to TLA, in the brain that might explain the discrepancies in the dopamine activations of the brain and the TD-error after learning \citep{masset2023multi}. 

We can perceive TLA as a novel control framework where three agents control the same body. This is different from traditional multi-agent approaches where each agent generally controls a different body or a different set of actuators. Often in multi-agent reinforcement learning, all the agents act at the same frequency and are often synced with the environmental time step. TLA offers an alternative approach where the agents act at different frequencies and thus are suited for different optimization goals for the same task. TLA uses time as a distinguishing factor to assign different goals and create a natural hierarchy between the agents. We plan to explore this multi-agent paradigm, especially in war games where agents act in different hierarchies and timescales.

The main limitation of TLA is that it can only plan a sequence of actions consisting of a single action. Therefore, the benefits of TLA are limited to multidimensional actions. In future work, we plan to implement a slow layer that can plan a sequence of actions instead of repeating the same action. In TLA, the fast layer acts as a reflex that only activates when needed, while the slow layer acts as a planning network in the brain. However, even within the brain, temporal attention is adaptable \citep{Morillon2016TemporalPI}. Thus, we will also explore making the step size of the slow layer adaptable to allow for changes in the planning horizon, while the fast network will remain for unexpected situations. This will be tested for improvements in the Walker2d environment.

Finally, beyond what was shown here, TLA may be most suitable for use in energy-constrained environments, environments that require a distributed approach, and environments with high communication costs or delays, such as drones and robotic systems. TLA sets a benchmark for decision and energy-constrained environments and paves the way for future research in time and energy-aware artificial intelligence.

\bibliographystyle{APA}
\bibliography{bibtex/bibliography}

\section*{Appendix}
\subsection*{Implementation details}
All experiments were performed on a GPU cluster with the following GPUs: Nvidia 1080ti, Nvidia TitanX, Nvidia 2080ti.

\subsection*{Hyperparameters}
The hyperparameters used for all our experiments for the TD3 algorithms are given below:
\begin{table}[h]
\centering
\begin{tabular}{|l|l|p{6.5cm}|}
    \hline
    Hyperparameter  &  Value & description \\
    \hline
    Exploration noise & 0.1 & Standard deviation of the Gaussian noise added to actions \\
    \hline 
    Batch size & 256 & Batch size for learning \\
    \hline 
    Discount factor & 0.99 & Discount factor \\
    \hline
    $\tau$ & 0.005 & Update rate for the target network in TD3 \\
    \hline
    Policy Noise & 0.2 & Noise added to target policy during critic update \\
    \hline
    Noise clip & 0.5 & Range to clip the target policy noise \\
    \hline
    Policy frequency & 2 & Delay factor for policy update \\
    \hline
    Replay Buffer Size & 1e6 & Size of the replay buffer\\ 
    \hline 
    
\end{tabular}
\caption{List of Common hyperparameters }
\label{appendixtable1}
\end{table}

\begin{table}[H]
\centering
\begin{tabular}{|l|l|l|l|l|l|l|}
    \hline
    Environment  &  $j$ & $p$ & $\tau$ & Start t & max T & eval frequency \\
    \hline
    Pendulum-v1 & 1 & 1 & 6 & 1000 & 30000 & 250 \\
    \hline 
   MountainCarContinuous-v0 & 1 & 1 & 11 & 10000 & 100000 & 2500 \\
    \hline 
     InvertedPendulum-v2 & 0.5 & 0.5 & 10 & 1000 & 1000000 & 5000 \\
    \hline 
    InvertedDoublePendulum-v2 & 4.5 & 4.5 & 5 & 1000 & 1000000 & 5000 \\
    \hline 
    Hopper-v2 & 0 & 1 & 9 & 20000 & 1000000 & 5000 \\
    \hline 
    Walker2d-v2 & 0 & 0.5 & 7 & 20000 & 1000000 & 5000 \\
    \hline 
    Ant-v2 & 0 & 0.2 & 3 & 20000 & 1000000 & 5000 \\
    \hline 
    HalfCheetah-v2 & 0 & 4.5 & 3 & 20000 & 1000000 & 5000 \\
    \hline
\end{tabular}
\caption{List of environment-specific hyperparameters }
\label{appendixtable1}
\end{table}

\newpage
\subsection*{TLA Architecture}
Here we provide a figure with more detailed architecture of TLA.

\begin{figure}[h]
\centering
\includegraphics[width=1.0\linewidth]{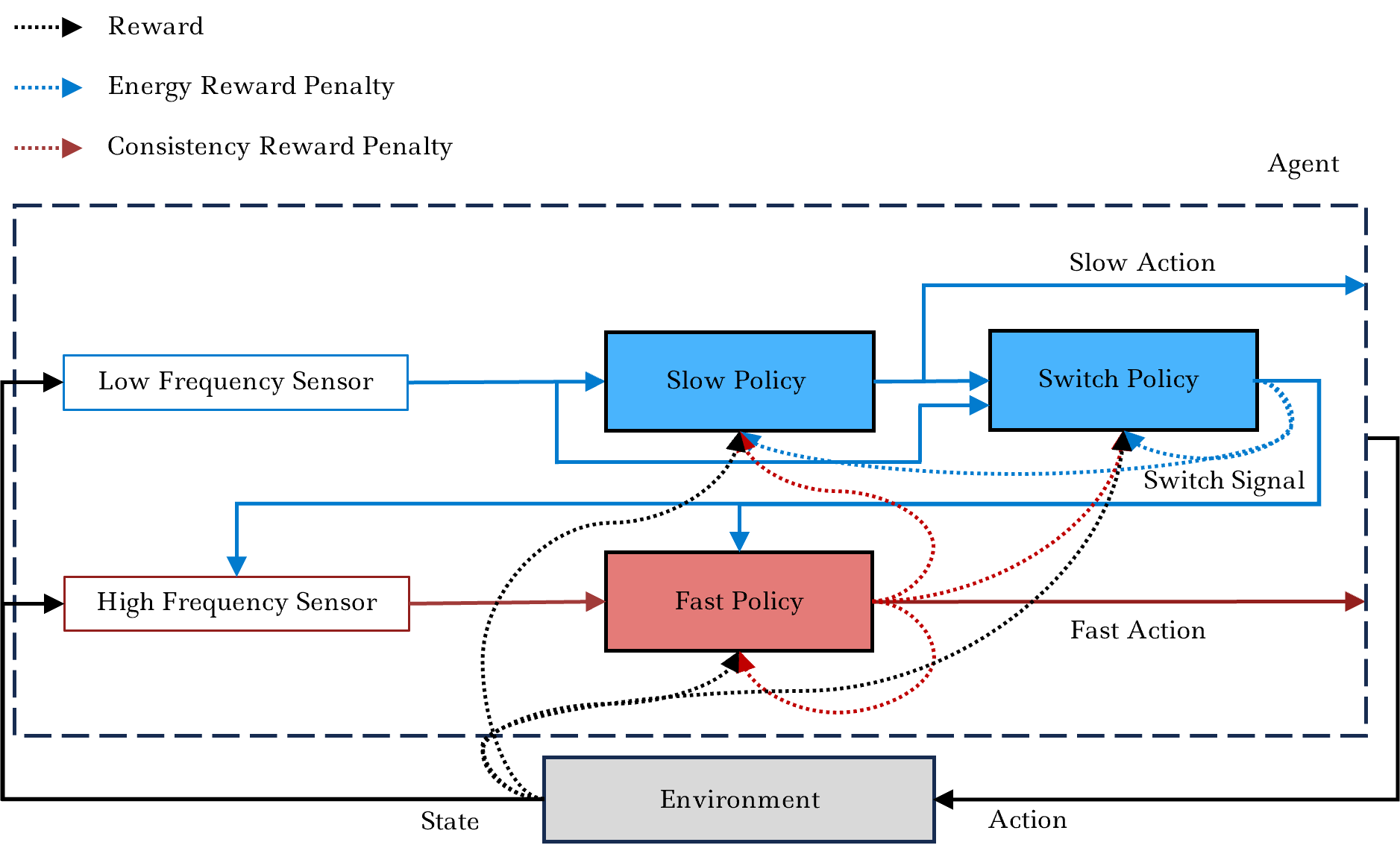}
\caption{The TLA architecture has two different layers: Slow (Blue) and Fast (Red). It has three different RL agents that are trained in parallel: Slow, Fast, and Switch. Each network receives a different reward penalty resulting in the optimization of energy in addition to the performance.}
\label{fig:TLA2}
\end{figure}
\newpage
\subsection*{TLA Algorithm}
\algsetup{
linenosize=\small,
linenodelimiter=.
}
\begin{algorithmic}
   \STATE Initialize Slow network with critics $Q_{\theta_{l1}}^l, Q_{\theta_{l2}}^l $ and actor $\pi_{\phi_l}^l$

  \STATE Initialize Fast network with critics $Q_{\theta_{q1}}^q, Q_{\theta_{q2}}^q $ and actor $\pi_{\phi_q}^q$

   \STATE Initialize Gate network with Q-function $Q_{\theta}^g$

   \STATE Initialize target networks $\theta_{l1}' \leftarrow \theta_{l1}, \theta_{l2}' \leftarrow \theta_{l2}, \phi_l' \leftarrow \phi_l, \theta_{q1}' \leftarrow \theta_{q1}, \theta_{q2}' \leftarrow \theta_{q2}, \phi_q' \leftarrow \phi_q$
   \STATE Initialize replay buffers $\mathcal{B}_l, \mathcal{B}_q, \mathcal{B}_g$
   \FOR{$t=1$ \textbf{to} $T$}
        \IF{$t$  mod $\tau$ }
         \STATE select slow action with exploration noise $a_l \sim \pi^l(s)+\epsilon,  \epsilon \sim \mathcal{N}(0,\sigma)$
         \STATE select gate action with $epsilon$ greedy policy $a_g \sim \pi^g(s, a_l)$
         \IF{$a_g$ = 0 }
         \STATE $r_l = 0$
         \STATE $r_g = 0$
         \STATE $s_l = 0$
         \ELSE
            \STATE $r_l = -p * \tau$
            \STATE $r_g =  -j * \tau$
         \ENDIF
        \ENDIF
        \IF{$a_g$ = 1 }
        \STATE  select fast action with exploration noise $a_q \sim \pi^q(s)+\epsilon, \epsilon \sim \mathcal{N}(0,\sigma)$
        \STATE $a = a_q$
        \ELSE
        \STATE $a = a_l$
        \ENDIF
        \STATE Perform the action $a$ and observe the reward $r$ and new state $s'$
        \IF{$a_g$ = 1 }
        \STATE  $r = r - j(a/a_{max})$
        \ENDIF
        \STATE Store transition tuple $(s,a,r,s')$ in $\mathcal{B}_q$
        \STATE $r_p += r$
        \STATE $r_g += r$
        \IF{$t+1$  mod $\tau$ }
        \STATE Store transition tuple $(s_l,a_l,r_l,s')$ in $\mathcal{B}_l$
        \STATE Store transition tuple $((s_l, a_l), a_g, r_g,s')$ in $\mathcal{B}_g$
        \ENDIF
        \STATE Sample mini-batches and update parameters for slow and fast according to TD3
        \STATE  Sample mini-batch and update parameters for Gate according to Q-learning
  \ENDFOR
\end{algorithmic}

\newpage
\subsection*{Detailed Results}

Here we provide the detailed results of our continuous-control experiments.

\begin{table}[H]
\centering
\resizebox{\columnwidth}{!}{
\begin{tabular}{|l|l|l|l|l||l|l|l|}
    \hline
    Environment & $\tau$ &\multicolumn{3}{|c|}{Normalized AUC}  & \multicolumn{3}{|c|}{average Return} \\
    
    \hline \hline
    && TD3 & TempoRL & TLA  & TD3 & TempoRL & TLA\\
    \hline \hline
    Pendulum &6& 0.85 & 0.85 & 0.87 & -147.38 (29.68) & -149.38 (44.64) & -154.92 (31.97)  \\
    \hline
   MountainCar &11& 0.19 & 0.64 & 0.82  & 0(0) & 84.56(28.27) & 93.88 (0.75) \\
   \hline
   Inv-Pendulum &10& 0.97 & 0.77 & 0.96 & 1000 (0) & 984.21 (47.37) & 1000(0) \\
   \hline
   Inv-DPendulum &5& 0.96 & 0.94 & 0.92  & 9359.82(0.07) & 9352.61(2.20) & 9356.67 (1.23) \\
   \hline
    Hopper & 9 & 0.66 & 0.43 & 0.75 & 3439.12 (120.98) & 2607.86 (342.23) & 3458.22 (117.92) \\
   \hline
   Walker2d & 7 &  0.56  & 0.52 & 0.53 & 4223.47 (543.6) & 4581.69 (561.95) & 3878.41 (493.97)  \\
   \hline
   Ant & 3 &  0.6 & 0.33 & 0.52  & 5131.90(687.00) & 3507.85 (579.95) & 5163.54 (573.19)  \\
   \hline
   HalfCheetah & 3 & 0.79 & 0.5 & 0.58  & 10352.58 (947.69) & 6627.74 (2500.78) & 9571.99 (1816.02)  \\
   \hline
   
\end{tabular}
}
\caption{Average normalized Area-under-curve (AUC) and average return results. The standard deviation is reported in parentheses. All results are averaged over 10 trials.}
\label{table:auc}
\end{table}

\begin{table}[H]
\centering
\resizebox{\columnwidth}{!}{
\begin{tabular}{|l|l|l|l||l|l|l|l|}
    \hline
    Environment & $\tau$ &\multicolumn{3}{|c|}{Action repetition}  & \multicolumn{3}{|c|}{average Jerk/time step} \\
    
    \hline \hline
    && TD3 & TempoRL & TLA  & TD3 & TempoRL & TLA\\
    \hline \hline
    Pendulum &6& 7.44$\%$ & 34.94$\%$ & 70.32$\%$ & 1.02 & 0.94 & 0.62  \\
    \hline
   MountainCar &11& 9.08$\%$ & 75.99$\%$ & 91.4$\%$  & 0.1 & 1.12 & 1.11 \\
   \hline
   Inv-Pendulum &10& 1.12$\%$ & 45.97$\%$ & 88.82$\%$ & 1.11 & 1.62 & 0.12 \\
   \hline
   Inv-DPendulum &5& 0.95$\%$ & 14.9$\%$ & 75.22$\%$  & 0.1 & 0.61 & 0.14 \\
   \hline
    Hopper & 9 & 2.51$\%$ & 64.99$\%$ & 57.22$\%$ & 0.46 & 0.4 & 0.25 \\
   \hline
   Walker2d & 7 &  2.14$\%$ & 69.47$\%$ & 47.45$\%$ & 0.27 & 0.2 & 0.21  \\
   \hline
   Ant & 3 &  0.82$\%$ & 22.01$\%$ & 12.68$\%$  & 0.43 & 0.39 & 0.38  \\
   \hline
   HalfCheetah & 3 & 5.64$\%$ & 14.07$\%$ & 18.05$\%$  & 0.8 & 0.65 & 0.67  \\
   \hline
   
\end{tabular}
}

\caption{Average action repetition percentage and jerk per time step. All results are averaged over 10 trials. Action repetition is measured as the percentage of actions that are the same as the previous action taken.}
\label{table:ar}
\end{table}

\begin{table}[H]
\centering
\resizebox{\columnwidth}{!}{
\begin{tabular}{|l|l|l|l|l|l|l|}
    \hline
    Environment &\multicolumn{3}{|c|}{average Decisions}  & \multicolumn{3}{|c|}{average MMACs}\\
    
    \hline \hline
    & TD3 & TempoRL & TLA  & TD3 & TempoRL & TLA \\
    \hline \hline
    Pendulum & 200 & 139.39 & 62.31 & 24.30 & 34.14 & 12.42
 \\
    \hline
   MountainCar & 999 & 116.47 & 10.6  & 120.98 & 28.60 & 2.54 \\
   \hline
   Inv-Pendulum & 1000 & 532.57 & 111.79 & 121.90 & 131.01 & 26.05\\
   \hline
   Inv-DPendulum & 1000 & 850.95 & 247.76  & 124.70 & 213.59 & 57.46 \\
   \hline
    Hopper & 998.99 & 269.85 & 423.91 & 125.17 & 68.43 & 72.02 \\
   \hline
   Walker2d &  988.17 & 297.4 & 513.12 & 127.08 & 77.29 & 92.07  \\
   \hline
   Ant  &  960.57 & 741.22 & 860.21  & 160.22 & 248.53 & 243.22  \\
   \hline
   HalfCheetah & 1000 & 889.57 & 831.42  & 128.60 & 230.13 & 182.35 \\
   \hline
   
\end{tabular}
}
\caption{Average decisions and million multiply-accumulate operations (MMACs) per episode and the multi-objective score for all environments. Decisions and MMACs are averaged over ten trials.}
\label{table:decisions}
\end{table}



\end{document}